\documentclass[letterpaper, 10pt, conference]{ieeeconf}

\makeatletter
\let\NAT@parse\undefined%
\makeatother

\IEEEoverridecommandlockouts%
\usepackage{mathtools}
\usepackage{amssymb}
\usepackage{booktabs} 
\usepackage{bm} 
\usepackage[usenames,dvipsnames]{xcolor}
\usepackage{graphicx}
\usepackage{balance} 
\DeclarePairedDelimiterX{\norm}[1]{\lVert}{\rVert}{#1}
\DeclarePairedDelimiterX\set[1]\lbrace\rbrace{#1}
\usepackage{nicefrac}
\usepackage[colorlinks,unicode,breaklinks]{hyperref} 
\hypersetup{
	pdftitle    = {Contact-Implicit Trajectory Optimization using an Analytically Solvable Contact Model for Locomotion on Variable Ground},
	pdfauthor   = {Iordanis Chatzinikolaidis, Yangwei You, and Zhibin Li},
    pdfsubject  = {Robotics},
    pdfkeywords = {Optimization and Optimal Control; Multi-Contact Whole-Body Motion Planning and Control; Contact Modeling; Legged Robots},
    breaklinks  = true,
    colorlinks  = true,
	urlcolor    = Blue,
	linkcolor   = Blue,
	citecolor   = Blue
}
\usepackage{subcaption}
\usepackage[capitalise,nameinlink]{cleveref}
\crefname{section}{Sec.}{Secs.}
\crefname{appsec}{Appendix}{Appendices}
\let\oldnamecref\namecref%
\renewcommand{\namecref}[1]{\hyperref[#1]{\oldnamecref{#1}}}
\usepackage{array}
\usepackage{tabularx}
\newcolumntype{M}[1]{>{\centering\arraybackslash}m{#1}}
\usepackage[all]{hypcap} 
\usepackage{etoolbox}
\apptocmd{\sloppy}{\hbadness 10000\relax}{}{}
\usepackage{xspace} 
\newcommand{\eg}{\hbox{\emph{e.g.}}\xspace}
\newcommand{\ie}{\hbox{\emph{i.e.}}\xspace}
\newcommand{\etc}{\hbox{\emph{etc.}}\xspace}
\usepackage[utf8]{inputenc}\DeclareUnicodeCharacter{2212}{-}
\DeclareMathOperator*{\argmin}{arg\,min}
\DeclareMathOperator{\smax}{smax}

\usepackage{tikz}
\newcommand\copyrighttext{%
  \footnotesize \textcopyright 2020 IEEE. Personal use of this material is permitted.
  Permission from IEEE must be obtained for all other uses, in any current or future media, including reprinting/republishing this material for advertising or promotional
  purposes, creating new collective works, for resale or redistribution to servers or lists, or reuse of any copyrighted component of this work in other works.
  DOI: \href{https://doi.org/10.1109/LRA.2020.3010754}{10.1109/LRA.2020.3010754}}
\newcommand\copyrightnotice{%
\begin{tikzpicture}[remember picture,overlay]
  \node[anchor=south,yshift=10pt] at (current page.south) {\fbox{\parbox{\dimexpr\textwidth-\fboxsep-\fboxrule\relax}{\copyrighttext}}};
\end{tikzpicture}%
}

\title{\LARGE \bf
Contact-Implicit Trajectory Optimization using an Analytically Solvable Contact Model for Locomotion on Variable Ground
}
\author{Iordanis Chatzinikolaidis\(^1\), Yangwei You\(^2\), and Zhibin Li\(^1\)%
\thanks{This research is supported by the EPSRC as part of the CDT in Robotics and Autonomous Systems (EP/L016834/1) and the EPSRC UK RAI Hub in Offshore Robotics for Certification of Assets (ORCA) (EP/R026173/1).}
\thanks{\(^1\)Authors are with the Edinburgh Centre for Robotics and School of Informatics, The University of Edinburgh, Edinburgh, UK\@.
Corresponding authors' emails: \texttt{\footnotesize iordanis.cs@gmail.com}, \texttt{\footnotesize zhibin.li@ed.ac.uk}.}
\thanks{\(^2\)Author is with the Institute for Infocomm Research, Agency for Science, Technology and Research (A*STAR), Singapore.}%
}

\begin{document}

\maketitle
\copyrightnotice
\thispagestyle{empty}
\pagestyle{empty}

\begin{abstract}
This paper presents a novel contact-implicit trajectory optimization method using an analytically solvable contact model to enable planning of interactions with hard, soft, and slippery environments.
Specifically, we propose a novel contact model that can be computed in closed-form, satisfies friction cone constraints and can be embedded into direct trajectory optimization frameworks without complementarity constraints.
The closed-form solution decouples the computation of the contact forces from other actuation forces and this property is used to formulate a minimal direct optimization problem expressed with configuration variables only.
Our simulation study demonstrates the advantages over the rigid contact model and a trajectory optimization approach based on complementarity constraints.
The proposed model enables physics-based optimization for a wide range of interactions with hard, slippery, and soft grounds in a unified manner expressed by two parameters only.
By computing trotting and jumping motions for a quadruped robot, the proposed optimization demonstrates the versatility for multi-contact motion planning on surfaces with different physical properties.
\end{abstract}

\section{Introduction}

Physical interactions in humans, animals, and robots require models of the contact properties in the environment to plan movements.
The necessary contact forces are generated due to intricate interactions between the contact media and are in practice difficult to model.
Therefore, proper modeling of these interactions is important for a motion planning framework that aims to produce contact-rich behaviors on variable grounds, such as locomotion on soft floors.

\begin{figure}
	\centering
	\hfill
	\subcaptionbox*{\bf Slippery ground}{\vspace{-5.5mm}\includegraphics[width=0.48\linewidth]{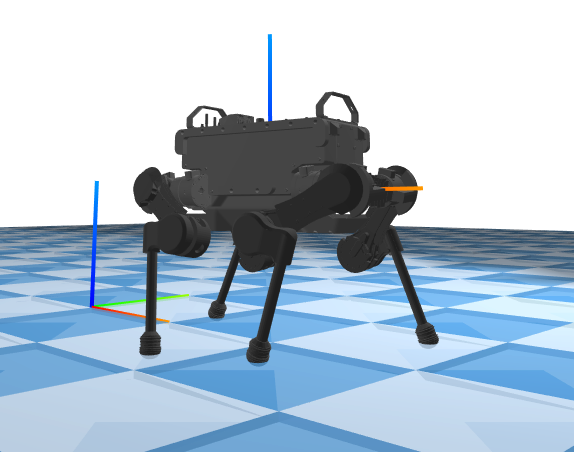}}
	\hfill
	\subcaptionbox*{\bf Soft ground}{\vspace{-5.5mm}\includegraphics[width=0.48\linewidth]{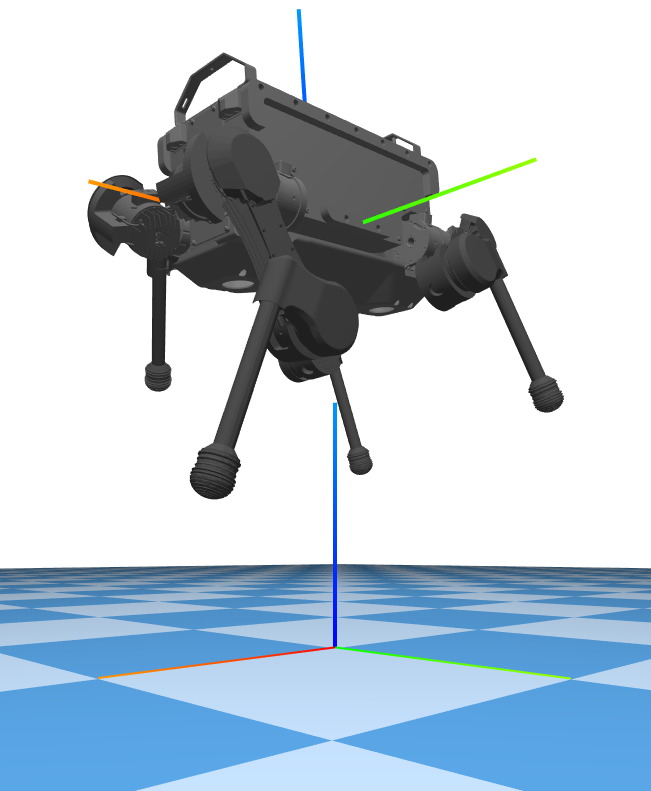}}
	\hfill
	\caption{Dynamic motions computed by the proposed framework: trotting on slippery ground (left); jumping on soft ground (right). (See slippery, soft ground effects in video).}\label{fig:teaser}
	\vspace{11mm}
\end{figure}

Environments in our daily life exhibit many properties: they can be hard, soft, slippery, or combinations thereof, as shown in \cref{fig:teaser}.
In terms of modeling, some of these aspects are usually missing in typical motion planning because most contact interactions are assumed rigid.
For example, two usual approaches are using either spring-damper~\cite{fahmi2020} or ad hoc penalization schemes~\cite{carius2019}.
In this work, we present a novel contact model in a principled formalism that can capture such properties without the drawbacks of spring-damper models.

A common solution in complex multi-contact planning is to split the original problem in a series of stages, obtaining more tractable subproblems that are still able to solve the original one.
Examples are the work in~\cite{tonneau2018} for general contact plans or the work in~\cite{carpentier2018a} with a pre-specified contact sequence.
Their main benefit is fast computation since each stage is usually designed to be efficiently solvable.
However, it is challenging to properly design these stages to compute general plans, without restricting the solution space or leading to infeasibilities for the subsequent stages.
The focus here is on approaches that avoid such decompositions and can reason about the generated motion plans holistically.

Trajectory optimization (TO) has emerged as a powerful framework to design locally optimal trajectories for highly dynamical and underactuated systems~\cite{kelly2017,betts2010}.
One of its main benefits is that it allows the setting of high-level goals, expressed as cost functions, while outputting a variety of motions as solutions.
This is especially important for legged locomotion and multi-contact motion.
Traditional approaches struggle to generalize across different scenarios or non-periodic motions, while TO methods are significantly more versatile~\cite{neunert2017a,radulescu2017}.
Expressing complex multi-contact planning through a TO lens enables the solution of a wide range of problems with minimum modifications.
Though learning approaches demonstrate similar generalization~\cite{yang2020,dallali2012}, this letter focuses on the model-based optimization paradigm.

A significant problem in TO with contacts is proper modeling as planning requires discontinuous and combinatorial reasoning.
Thus, some approaches focus on embedding part of the problem in the TO description.
This is possible for simple legged systems; for instance, one-leg hoppers~\cite{chatzinikolaidis2018} and bipeds with a pre-defined periodic gait pattern like running~\cite{mombaur2009}.
While this leads to problems with a very specific structure that are usually easier to solve, adapting to different legged configurations is difficult.
This work focuses on approaches that work for arbitrary legged systems.
This is possible for contact-implicit formulations that do not require a priori specification of the contact sequence.

An alternative approach is to describe the problem in a bilevel fashion: The outer level updates the state of the model, while the inner level computes the contact information~\cite{yunt2006,carius2018,erez2012}.
Bilevel methods are usually solved by formulating the Karush–Kuhn–Tucker conditions of the inner level, leading to a mathematical program with complementarity constraints (MPCC)~\cite{dempe2012}.
Contrary to bilevel approaches, our work focuses on direct methods that formulate and solve the problem in a single level, such that the optimizer can reason about contacts by optimizing forces, with benefits for long-term physical reasoning~\cite{toussaint2020}.

Enabling the optimizer to directly reason about contact forces was proven very powerful for generating complicated contact-implicit motion plans.
One way is to allow contact forces to act from a distance~\cite{mordatch2012}; while this is important for discovering contacts, penalizing these forces for physically realistic motion can be challenging.
A more principled formalism is introduced in~\cite{posa2014}, where the problem is elegantly posed as an MPCC\@.
This allows leveraging relaxations for this class of problems already studied in the optimization literature.
Albeit these relaxations, a fundamental problem lies in the complementarity constraints, which usually violate constraint qualification tests~\cite{betts2010}.
Some of these constraints are due to the contact model used.

Therefore, our principal motivation is to introduce a contact model that does not require the specification of complementarity constraints.
Such an idea is discussed in~\cite{drumwright2011}, where they propose a pair of convex optimization problems that compute the contact forces for simulation purposes.
For direct TO, the previous work either focused on MPCC formulations or used a spring-damper model~\cite{neunert2017a,onol2019}.

In this work, we present a TO formulation with a contact model expressed as a pair of quadratic problems that can be computed in closed form.
Thus, complementarity constraints are not required while problems associated with spring-damper models such as energy injections, stiff differential equations, and difficulties imposing the friction cone constraint are mitigated.
Furthermore, our framework allows deriving the equations of motions for physical interaction with environments characterized by different stiffness, viscosity, and friction.
By using the proposed framework, a variety of motion plans can be computed for hard, soft, and slippery surfaces by setting a small number of parameters.
The contributions are summarized as follows:
\begin{itemize}
	\item An analytically solvable contact model suitable for direct contact-implicit TO\@, which can be utilized in formulations without complementarity constraints, while satisfying unilaterality and friction cone constraints.

	\item The proposed contact model is generic and can be used to compute motion plans on hard, soft, and slippery surfaces in a unified manner.

	\item A TO framework that integrates the new contact model for generating contact-implicit motion plans for a high degree of freedom robot, demonstrating the advantages of the proposed method with extensive comparisons performed against the rigid contact model and a TO formulation with complementarity constraints.
\end{itemize}

The remaining sections are organized as follows.
\cref{sec:background} describes how contacts are resolved in a simulation setting and introduces two contact models.
In \cref{sec:TO_formulation}, the proposed contact model is derived and the overall direct TO formulation is elaborated.
\cref{sec:experiments} presents the comparisons with (i) a different contact model, (ii) an alternative direct TO formulation based on complementarity constraints, (iii) followed by a variety of computed quadrupedal motions on terrains with different properties.
Finally, we summarize and discuss future outlooks in \cref{sec:conclusion}.

\section{Background and Preliminaries}\label{sec:background}

\subsection{Dynamics with contacts}

The equations of motion of a typical robot model are
\begin{equation}\label{eq:langrangian_dynamics}
	M(\bm{q})\ddot{\bm{q}} + \bm{H}(\bm{q}, \dot{\bm{q}}) = S\bm{\tau} + \sum_i J_i^T(\bm{q})\bm{f}_i,
\end{equation}
where \(M\) is the mass matrix, \(\bm{H}\) the vector of nonlinear forces (Coriolis, centrifugal, and gravitational), \(S\) is a selection matrix that maps actuated joint torques \(\bm{\tau}\) to generalized coordinates, while \(J_i\) denotes the Jacobian of the \(i\)-th contact and \(\bm{f}_i\) the corresponding force.
We simplify the notation by dropping explicit dependence on quantities and write \(M\) instead of \(M(\bm{q})\), \etc
Furthermore, we denote the generalized velocity and acceleration as the time derivative of the configuration, though this is not necessarily the case; for example, if the floating base is represented using quaternions, the angular velocity of the base is not equal to the rate of change of the quaternions.

Next, we follow a similar treatment with time-stepping approaches, \eg~\cite{todorov2014,horak2019}.
Equation~\eqref{eq:langrangian_dynamics} is discretized using an Euler approximation to obtain the discrete version
\begin{equation}
	M_k\left(\dot{\bm{q}}_{k+1} - \dot{\bm{q}}_k\right) = h(S\bm{\tau}_k - \bm{H}_k) + J_k^T \bm{\lambda}_k,
\end{equation}
where \(h\) is the time step and \(\bm{\lambda}_k\) corresponds to the concatenation of the contact impulses at time step \(k\).
Next, the discrete equations of motions are projected in contact space
\begin{equation}\label{eq:proj_eom_full}
	J_k \left(\dot{\bm{q}}_{k+1} - \dot{\bm{q}}_k\right) = J_k M^{-1}_k \left[h (S\bm{\tau}_k - \bm{H}_k) + J_k^T \bm{\lambda}_k \right].
\end{equation}
Alternatively,~\eqref{eq:proj_eom_full} can be expressed as
\begin{equation}\label{eq:proj_eom}
	\bm{v}^+ = A \bm{\lambda} + \bm{b}(\bm{\tau}) + \bm{v}^-,
\end{equation}
with \(\bm{v}^+ = J_k \dot{\bm{q}}_{k+1}\), \(\bm{v}^- = J_k \dot{\bm{q}}_k\), \(\bm{b} = h J_k M^{-1}_k (S\bm{\tau}_k - \bm{H}_k)\), and \(A = J_k M^{-1}_k J_k^T\).

Based on~\eqref{eq:proj_eom}, two cases are identified~\cite{todorov2014}: (i) The \textit{forward contact dynamics} case, where we want to compute \(\left(\bm{v}^+, \bm{\lambda}\right)\) given \(\left(A, \bm{b}, \bm{v}^-\right)\); (ii) The \textit{inverse contact dynamics} case, where we want to compute \(\left(\bm{b}, \bm{\lambda}\right)\) given \(\left(A, \bm{v}^-, \bm{v}^+\right)\).
The forward contact dynamics is more relevant in simulation, while the inverse is more relevant in a TO setting.
In the latter case, decomposing actuation from contact impulses for an underactuated robot model is challenging, which is an aspect addressed in this work.

Finally, the contact impulses should satisfy unilateral and friction cone constraints, \ie each impulse must satisfy
\begin{equation}\label{eq:contact_impulse_set}
	F_{\mu_i} = \big \{ \,\bm{\lambda}_i \mid \lambda_{n(i)} \geq 0, \norm{\bm{\lambda}_{t(i)}}^2 \leq \mu_i^2 \lambda^2_{n(i)}\,\big \} ,
\end{equation}
where \(\bm{\lambda}_i = \begin{bmatrix} \bm{\lambda}_{t(i)} & \lambda_{n(i)} \end{bmatrix}^T\) are the tangential and normal components.
Equations~\eqref{eq:proj_eom} and~\eqref{eq:contact_impulse_set} form the backbone of impulse-based time-stepping methods~\cite{stewart2000}.
The contact models discussed next provide different approaches on how to solve them.

\subsection{Contact models}

\textit{Nonlinear complementarity problem (NCP)}.
This contact model augments~\eqref{eq:proj_eom} and~\eqref{eq:contact_impulse_set} with an additional constraint:
That either the contact normal distance or the normal contact impulse at the next time instant is zero.
This is usually described succinctly as \(0 \leq \lambda_n \perp d^+ \geq 0\), where \(d^+\) is the next step normal gap distance required to be nonnegative to avoid penetration.
In simulation \(d^+\) is generally unknown, so this condition is approximated by performing its Taylor expansion and forming a complementarity constraint between the gap velocity and normal contact impulse~\cite{horak2019}.
Apart from the friction cone constraint, the tangential components of the impulse are specified via the \textit{maximum dissipation principle} (MDP).
It specifies that friction forces maximize the rate of the kinetic energy dissipation.
This principle was introduced in~\cite{moreau2011} for a frictional contact at a single point.

\textit{Convex optimization formulation}.
The convex contact model relaxes the complementarity condition by forming a quadratic problem that penalizes movement in contact space~\cite{todorov2014}.
In the forward dynamics case, it is specified by the quadratically constrained quadratic program~\cite{horak2019}
\begin{equation}\label{eq:fd_convex_model}
	\begin{IEEEeqnarraybox}[][c]{c'c}
		\underset{\bm{\lambda}}{\text{min}} & \frac{1}{2} \bm{\lambda}^T (A + R) \bm{\lambda} + \bm{\lambda}^T (\bm{b} + \bm{v}^-)\\
		\text{s.t.} & \bm{\lambda}_i \in \mathbb{F}_{\mu_i} \text{, } \forall i
	\end{IEEEeqnarraybox}
\end{equation}
that computes the contact impulses, where \(R\) is a positive definite matrix that makes the solution unique and invertible.
Physically this makes the contact model soft; for hard contacts, there can be an infinite number of contact impulse combinations.
The inverse dynamics case is well-defined and the impulses are given by
\begin{equation}\label{eq:id_convex_model}
	\begin{IEEEeqnarraybox}[][c]{c'c}
		\underset{\bm{\lambda}}{\text{min}} & \frac{1}{2} \bm{\lambda}^T R \bm{\lambda} + \bm{\lambda}^T \bm{v}^+ .\\
		\text{s.t.} & \bm{\lambda}_i \in \mathbb{F}_{\mu_i} \text{, } \forall i.
	\end{IEEEeqnarraybox}
\end{equation}
For a diagonal \(R\), the latter optimization problem splits into independent problems, one for each contact.

\section{Trajectory Optimization Formulation}\label{sec:TO_formulation}

\subsection{Optimal control problem}

The continuous optimal control problem (OCP) can be expressed as
\begin{IEEEeqnarray}{C'C}\label{eq:continuous_TO}
    \IEEEyesnumber*
    \IEEEyessubnumber*
    \min_{\bm{q}(t), \bm{\tau}(t)} & l_f(\bm{q}_T) + \int_0^T l(\bm{q}) + c(\bm{\tau}) dt \label{eq:TO_cost}\\
    \text{s.t.} & M(\bm{q})\ddot{\bm{q}} + \bm{H}(\bm{q}, \dot{\bm{q}}) = S\bm{\tau} +  J^T(\bm{q})\bm{f} \label{eq:TO_system_model}\\
     & \bm{f} \in \left\{\begin{IEEEeqnarraybox}[\IEEEeqnarraystrutmode\IEEEeqnarraystrutsizeadd{2pt}{2pt}][c]{CC}
        \argmin_{\bm{f}} & k(\bm{q}, \dot{\bm{q}}, \bm{f})\\
        \text{s.t.} & \bm{f} \in F_\mu
    \end{IEEEeqnarraybox}\right. \label{eq:TO_forces_opt}\\
     & \bm{g}\left(\bm{q}, \bm{\tau}\right) \in \mathbb{Z} \label{eq:TO_path_constraints}\\
     & \begin{IEEEeqnarraybox}[\IEEEeqnarraystrutmode\IEEEeqnarraystrutsizeadd{2pt}{2pt}][c]{C}
         \bm{q}(0) = \bm{q}^0\\
         \dot{\bm{q}}(0) = \dot{\bm{q}}^0
       \end{IEEEeqnarraybox}\label{eq:TO_initial_state}\\
     & t \in [0, T], \label{eq:TO_time}
\end{IEEEeqnarray}
where \(l(\bm{q})\) is an additive cost associated with the joint trajectory, \(l_f(\bm{q}_T)\) is the final state cost, and \(c(\bm{\tau})\) is the cost of the joint torques.
These can be general sufficiently smooth functions but we focus on positive definite quadratic forms.
Equation~\eqref{eq:TO_system_model} specifies the dynamics of the system, where \(J(\bm{q})\) and \(\bm{f}\) are the concatenated Jacobians and contact forces, respectively.
The constraints~\eqref{eq:TO_path_constraints} specify general path constraints imposed on the optimal trajectory, \eg joint and torque limits.
Finally,~\eqref{eq:TO_initial_state} and~\eqref{eq:TO_time} specifies the initial state and the time.

The optimization problem in~\eqref{eq:TO_forces_opt} specifies the contact forces via a mathematical program which makes~\eqref{eq:continuous_TO} a bilevel optimization problem.
The first approach is to provide gradients to the upper level via sensitivity analysis, but this can make long-term physical reasoning hard~\cite{toussaint2020}.
The second approach is to introduce the contact forces as variables, and then describe~\eqref{eq:TO_forces_opt} via its first-order necessary (KKT) conditions; that is, by imposing complementarity constraints.
For example, the NCP contact model requires constraints for avoiding penetrations and the KKT conditions of the MDP~\cite{manchester2019}.
A third approach is to solve~\eqref{eq:TO_forces_opt} for the contact forces, which are then not introduced as variables, and the associated complementarity constraints become unnecessary.
We present such an approach next.

\subsection{Contact model with analytical solution}\label{sec:contact_model}

First, the solution for the normal components is specified.
We call it the \textit{frictionless case}.
To obtain a unique solution, the strict non-penetration constraint is replaced with a quadratic program that has a unique solution and penalizes penetrations and the magnitude of the normal forces, while satisfying unilateral contact impulse constraints.

Given this solution, the tangential components are computed for what we call the \textit{friction case}. Instead of the MDP, the velocities in contact space are minimized as in~\cite{todorov2014}.
Since we focus on TO, the advantage of this approach is that an invertible contact model for the tangential components analogous to~\eqref{eq:id_convex_model} can be formulated, and its unique minimum can be analytically derived.

\subsubsection{Frictionless case}

The following quadratic problem specifies each normal component as
\begin{equation}\label{eq:frictionless_opti}
	\begin{IEEEeqnarraybox}[][c]{c'c}
		\underset{\lambda_{n(i)}}{\text{min}} & \frac{1}{2r_{n(i)}} \lambda^2_{n(i)} + \lambda_{n(i)} d^+_i(q) .\\
		\text{s.t.} & \lambda_{n(i)} \geq 0.
	\end{IEEEeqnarraybox}
\end{equation}
Since the next step gap distance is available, the velocity complementarity constraint becomes unnecessary.
The solution is
\begin{equation}\label{eq:normal_solution}
	\lambda_{n(i)} = r_{n(i)} \max{\left \{-d^+_i(q), 0\right \}}.
\end{equation}
Notice that the model depends on information from the next time instant, in contrast to traditional spring models.
By examining the solution, penetration occurs when the contact impulses are activated and become positive; otherwise, they are zero.
Furthermore, \(r_{n(i)}\) expresses the trade-off between the magnitude of the normal impulse and the penetration depth.
The larger \(r_{n(i)}\), the smaller the penetration.

\subsubsection{Friction case}

The optimization problem that minimizes the velocities in the contact frame for the inverse dynamics case has the form
\begin{equation}\label{eq:friction_opti}
	\begin{IEEEeqnarraybox}[][c]{c'c}
		\underset{\bm{\lambda}_{t(i)}}{\text{min}} & \frac{1}{2} \bm{\lambda}_{t(i)}^T R_{t(i)}^{-1} \bm{\lambda}_{t(i)} + \bm{\lambda}_{t(i)}^T \bm{v}^+_{t(i)} \\
		\text{s.t.} & \norm{\bm{\lambda}_{t(i)}}^2 \leq \mu_i^2\lambda_{n(i)}^2,
	\end{IEEEeqnarraybox}
\end{equation}
where \( R_{t(i)} = \text{diag}\left \{r_{t(i)} \text{, } r_{t(i)} \right \} \).
Essentially, \(r_t\) trades off tangential velocity and force; the smaller its value, the more tangential forces are penalized.

This is a projection to circle problem and two cases can be identified.
If the solution lies inside the cone, the problem is an unconstrained quadratic one, and the solution is
	\begin{equation}\label{eq:friction_opti_sol1}
		\bm{\lambda}_{t(i)} = -R_{t(i)} \bm{v}^+_{t(i)}.
	\end{equation}
Otherwise, the solution lies on the boundary
	\begin{equation}\label{eq:friction_opti_sol2}
		\bm{\lambda}_{t(i)} = -\mu_i \lambda_{n(i)} \hat{\bm{v}}^+_{t(i)},
	\end{equation}
	where \(\hat{\bm{v}}^+_{t(i)} = \nicefrac{\bm{v}^+_{t(i)}}{\norm{\bm{v}^+_{t(i)}}}\).

For \(r_{t(i)} \rightarrow 0\) the solution approaches the frictionless case, while for \(r_{t(i)} \gg 0\) energy dissipation is increased.
The tangential components of the impulse are given by~\eqref{eq:friction_opti_sol1} if \(r_{t(i)} \norm{\bm{v}^+_{t(i)}} \leq \mu_i \lambda_{n(i)}\); otherwise, they are given by~\eqref{eq:friction_opti_sol2}.
Finally, the tangential force is opposite to the tangential velocity; thus, the reaction force is dissipative.

\subsubsection{Smoothing}

In continuous optimization, smoothness of the objective and the constraints for at least the second derivative is required~\cite{nocedal2006}.
From the previous analysis, it is clear that the computed impulses contain switches that can cause problems for the optimizer. Therefore, a procedure to remove the discontinuities is discussed below.
First, the solution for the \textit{friction case} is expressed using a \texttt{max} function,
\begin{equation}\label{eq:tangential_solution}
	\bm{\lambda}_{t(i)} = \hat{\bm{v}}^+_{t(i)}\max{\left \{-\mu_i \lambda_{n(i)}, -r_{t(i)} \norm{\bm{v}^+_{t(i)}}\right \}}.
\end{equation}

As a result, a smooth approximation to the \texttt{max} function is required, which is now present in the solutions for both cases.
Multiple definitions for a smooth max function exist, and the selected \texttt{softmax} function is
\begin{equation}
	\smax(\alpha, \beta; \epsilon) = \frac{\alpha + \beta + \sqrt{{(\alpha - \beta)}^2 + \epsilon^2}}{2},
\end{equation}
where for \(\epsilon > 0 \rightarrow 0\) the approximation becomes stricter.

\subsection{Direct transcription}\label{subsec:overall_optimization}

\begin{table}
    \caption{Parameters for the unactuated rigid body models}\label{tab:unactuated_model_params}
    \centering
    \begin{tabularx}{\linewidth}{M{0.1\linewidth} M{0.12\linewidth} M{0.18\linewidth} M{0.18\linewidth} M{0.177\linewidth}}
        \toprule
        Model & Position \([m]\) & Orientation (MRP) & Body angular vel. \([\nicefrac{rad}{s}]\) & Body linear vel. \([\nicefrac{m}{s}]\)\\
        \midrule
        Ball & \(\begin{bsmallmatrix} 0.1 \\ -0.75 \\ 0.3 \end{bsmallmatrix}\) & \(\begin{bsmallmatrix} -0.1617 \\ 0.566 \\ -0.0809 \end{bsmallmatrix}\) & \(\begin{bsmallmatrix} -0.372 \\ 1.208 \\ -0.834 \end{bsmallmatrix}\) & \(\begin{bsmallmatrix} -1.379 \\ -1.386 \\ -0.743 \end{bsmallmatrix}\)\\
        Brick & \(\begin{bsmallmatrix} 0.1 \\ -0.75 \\ 1.7 \end{bsmallmatrix}\) & \(\begin{bsmallmatrix} 0 \\ 0 \\ 0 \end{bsmallmatrix}\) & \(\begin{bsmallmatrix} -1 \\ -0.2 \\ 0.126 \end{bsmallmatrix}\) & \(\begin{bsmallmatrix} 0 \\ 0 \\ 0 \end{bsmallmatrix}\)\\
        \bottomrule
    \end{tabularx}
\end{table}

According to the solutions~\eqref{eq:normal_solution} and~\eqref{eq:tangential_solution}, the contact impulses are described as a function of joint configurations and velocities, and the Jacobian can be used to map these quantities to the contact velocity in~\eqref{eq:tangential_solution}.
Afterwards, these terms can be substituted in~\eqref{eq:TO_system_model}, which becomes a function of the joint positions, velocities, and accelerations only.

Note that~\eqref{eq:continuous_TO} is an infinite-dimensional continuous problem that can be transcribed to a finite discrete one~\cite{betts2010}.
An implicit Euler discretization is selected due to its numerical properties, \eg it is an A-stable method.
Thus, problem~\eqref{eq:continuous_TO} can be expressed as
\begin{IEEEeqnarray}{C'C}
    \IEEEyesnumber*
    \IEEEyessubnumber*
    \min_{\bm{q}_k, \dot{\bm{q}}_k, \bm{\tau}_k} & \> \> \> \>  l_f (\bm{q}_N) + h\sum_{k=1}^{N-1} l_k (\bm{q}_k, \dot{\bm{q}}_k) + c_k (\bm{\tau}_k) \> \> \> \> \> \> \> \> \> \> \>  \> \> \> \> \> \> \> \label{eq:dis_dyn_obj}
\end{IEEEeqnarray}
\begin{IEEEeqnarray}[\vspace{-\baselineskip}]{C?rCl}
    \IEEEyessubnumber*
    \text{s.t.} & M_{k+1}(\dot{\bm{q}}_{k+1} - \dot{\bm{q}}_k) & = & h \left(S\bm{\tau}_{k+1} - \bm{H}_{k+1} \right. \nonumber\\
     & & & \> \left. + J^T_{k+1} \lambda(\bm{q}_{k+1}, \dot{\bm{q}}_{k+1})\right) \> \> \> \> \label{eq:dis_dyn_vel}\\
     & h\dot{\bm{q}}_{k+1} & = & \bm{q}_{k+1} - \bm{q}_k\label{eq:dis_dyn_config}\\
     & \bm{g}\left(\bm{q}_k, \bm{\tau}_k\right) & \in & \mathbb{Z}\\
     & \begin{IEEEeqnarraybox}[\IEEEeqnarraystrutmode\IEEEeqnarraystrutsizeadd{2pt}{2pt}][c]{C} \bm{q}_0 \\ \dot{\bm{q}}_0 \end{IEEEeqnarraybox} & 
       \begin{IEEEeqnarraybox}[\IEEEeqnarraystrutmode\IEEEeqnarraystrutsizeadd{2pt}{2pt}][c]{C} = \\ = \end{IEEEeqnarraybox} &
       \begin{IEEEeqnarraybox}[\IEEEeqnarraystrutmode\IEEEeqnarraystrutsizeadd{2pt}{2pt}][c]{C} \bm{q}^0 \\ \dot{\bm{q}}^0 \end{IEEEeqnarraybox}\\
     & k & \in & [0,N-1].
\end{IEEEeqnarray}

Since~\eqref{eq:dis_dyn_config} is linear to joint velocities, they can be removed from the optimization problem by substituting the right-hand side.
Similarly, joint torques are split into actuated \(\bm{\tau}\) and underactuated \(\bm{\tau}^u\) parts using~\eqref{eq:dis_dyn_vel}~\cite{erez2012}.
Then \(\bm{\tau}\) is given as a function of the joint configurations and can be substituted directly on the objective~\eqref{eq:dis_dyn_obj}.
Finally, the underactuated dynamics should be zero, yielding the overall trajectory optimization problem
\begin{IEEEeqnarray}{C'C}
    \IEEEyesnumber*
    \IEEEyessubnumber*
    \min_{\bm{q}_k} & l_f (\bm{q}_N) + h\sum_{k=1}^{N-1} l_k (\bm{q}_k) + c_k (\bm{\tau}(\bm{q}_k, \bm{q}_{k+1})) \hspace{0.5cm}
\end{IEEEeqnarray}
\begin{IEEEeqnarray}[\vspace{-\baselineskip}]{C?rCl}
    \IEEEyessubnumber*
    \text{s.t.} & M_{k+1}^u (\bm{q}_{k+1} - 2\bm{q}_k + \bm{q}_{k-1}) & = & h^2 (J^{uT}_{k+1} \lambda_{k+1} \nonumber\\
     & & & \> - \bm{H}^u_{k+1})\\
     & \bm{g}\left(\bm{q}_k, \bm{\tau}_k\right) & \in & \mathbb{Z}\\
     & \begin{IEEEeqnarraybox}[\IEEEeqnarraystrutmode\IEEEeqnarraystrutsizeadd{2pt}{2pt}][c]{C} \bm{q}_0 \\ \bm{q}_{-1} \end{IEEEeqnarraybox} & 
       \begin{IEEEeqnarraybox}[\IEEEeqnarraystrutmode\IEEEeqnarraystrutsizeadd{2pt}{2pt}][c]{C} = \\ = \end{IEEEeqnarraybox} &
       \begin{IEEEeqnarraybox}[\IEEEeqnarraystrutmode\IEEEeqnarraystrutsizeadd{2pt}{2pt}][c]{L} \bm{q}^0 \\ \bm{q}^0 - h\dot{\bm{q}}^0  \end{IEEEeqnarraybox}\\
     & k & \in & [0,N-1],
\end{IEEEeqnarray}
where \({(\cdot)}^u\) denotes the unactuated part of the quantity.
It is worth noting that the Hessian and Jacobian remain sparse.

Finally, the modified Rodrigues parameters (MRP) are selected~\cite{terzakis2018} to represent the floating base.
As any three-parameter orientation parameterization, it possesses a singularity.
For the MRP representation, this singularity is located after a full revolution.
That places the singularity as far as possible from the origin.
The polynomial expression and lack of unit norm constraints make this representation a suitable candidate for nonlinear optimization.

\section{Results and Validation}\label{sec:experiments}

\begin{figure}
	\centering
	\includegraphics{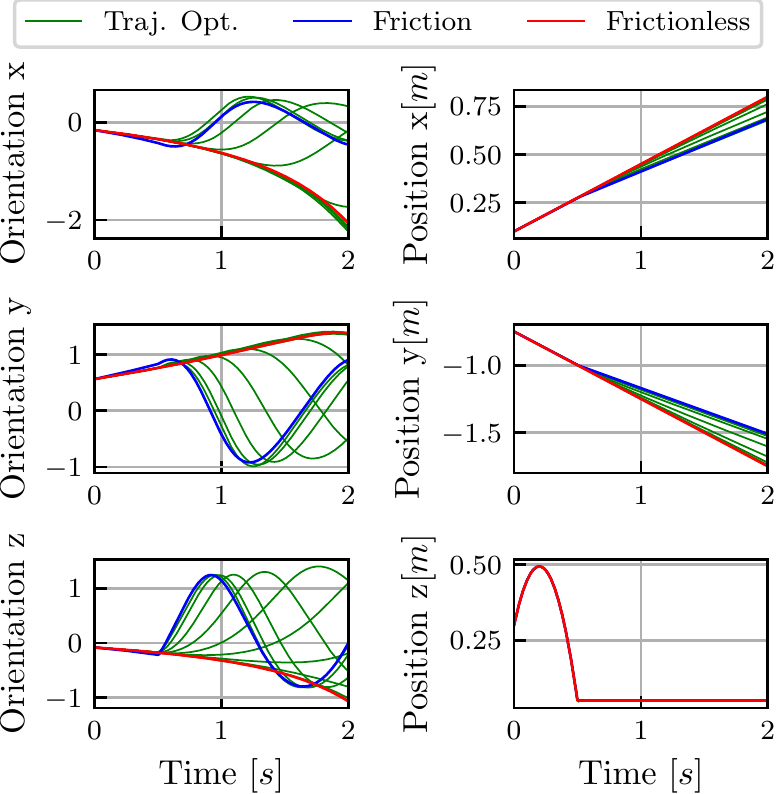}
	\caption{By changing the value of the parameter \(r_t\), we can obtain a family of solutions that range from the frictionless to the dissipative friction case.}\label{fig:to_vs_lcp_ball}
\end{figure}

Next, simulations are conducted to quantitatively validate the proposed formulation.
For all the cases, the optimization problems are formulated with \texttt{CasADi}~\cite{andersson2018} (for automatic differentiation), and solved by \texttt{IPOPT}, a large scale interior-point solver~\cite{wachter2006}.
\texttt{IPOPT} allows the selection of a linear solver for computing the Newton steps; we used the \texttt{MA57} solver when performing comparisons and \texttt{MA86} otherwise~\cite{hsl}.
The rigid body dynamics of the models are computed using the \texttt{RigidBodyDynamics.jl} library~\cite{koolen2019}.

As in all nonlinear optimization problems, proper scaling is important.
Since configurations are the only variables, scaling here is straightforward and only the position of the floating joint requires special treatment.
For the constraints and objective function, we use the default gradient-based scaling available in \texttt{IPOPT}.

First, the results for two basic models are studied before proceeding to a complex robot model: A rigid ball model that constitutes the simplest 3D floating model with one contact point and a rigid brick with eight contact points.
These models are suitable for benchmarking and require minimum parameter tuning; the small number of states and the unique state solution provide a framework for direct comparisons.
For benchmarking, we avoided more complicated models which can make the results less comparable due to the high-dimensional representations and non-trivial choices of parameters.

The initial state for all ball and brick simulations is summarized in~\cref{tab:unactuated_model_params}, where the root body spatial velocity is defined in the body frame, while we use \(\epsilon = 0.001\) for smoothing.
Regarding the simulated motions, the ball and the brick were dropped on a flat ground.
Parameter \(r_n\) is selected so that no bouncing occurs.
Thus, for the frictionless case, the ball slides on the surface, while it rolls when friction is present.
The dropped brick touches the plane with a line contact.
It slowly rotates until settling on a large side---where four vertices are active contact points\footnote{A video of all simulated cases is available at \href{https://youtu.be/eLx1DebDHmY}{youtu.be/eLx1DebDHmY}}.

\subsection{Comparison with a physics simulation}\label{subseq:sim_comparison}

\begin{table}
    \caption{Running time and iterations of the MPCC versus our proposed formulations}\label{tab:mpcc_our_comparison}
    \centering
    \begin{tabularx}{\linewidth}{M{0.17\linewidth}M{0.21\linewidth}M{0.06\linewidth}M{0.19\linewidth}M{0.06\linewidth} M{0.02\linewidth}}
        \toprule
        Method & \multicolumn{2}{c}{MPCC} & \multicolumn{2}{c}{Proposed}\\
               & Wall time \([s]\) & Iter. & Wall time \([s]\) & Iter. & n\\
        \midrule
        Ball frictionless & \(0.773 \pm 0.011\) & \(24\) & \(0.507 \pm 0.016\) & \(15\) & \(5\) \\
        Ball friction & \(3.78 \pm 0.047\) & \(104\) & \(2.374 \pm 0.023\) & \(64\) & \(5\) \\
        Brick frictionless & \(1340 \pm 15.27\) & \(6034\) & \(508.83 \pm 0.56\) & \(1893\) & \(3\) \\
        Brick friction & \(1151 \pm 20.18\) & \(3015\) & \(317.68 \pm 4.69\) & \(819\) & \(3\) \\
        \bottomrule
    \end{tabularx}
\end{table}

To evaluate the contact impulses computed by the proposed approach, we performed a comparison with the rigid contact model which is typically used in simulation engines.
The aim is to demonstrate that the model can represent different environmental interactions by an intuitive selection of its parameters.
Assuming that the frictionless case corresponds to an extremely slippery interaction, our model can simulate very slippery conditions (numerically identical to the frictionless one), up until minimally slippery (numerically identical to the friction case).

For the primitive models, the optimization is equivalent to a root-finding problem.
We perform comparisons against the nonlinear complementarity model with the Projected Gauss-Seidel (PGS) solver~\cite{horak2019}.
This implementation uses a semi-implicit Euler integration scheme for the dynamics (so that the problem remains linear) while we use an implicit one.
Furthermore, that trajectory is computed step by step since the computation is done in a simulation setting, whereas our TO formulation computes the whole trajectory simultaneously.

\cref{fig:to_vs_lcp_ball} illustrates the position and orientation of the ball.
First, the PGS solver was executed with and without friction and the resulting two solutions are plotted.
Afterwards, \(r_n = 100\nicefrac{N}{m}\) is fixed for the normal component, and a parameter sweep is performed for \(r_t\), with \(\mu = 0.5\).
It is verified that by changing the parameter \(r_t\), we obtain the friction and frictionless solutions, as well as additional in-between solutions; these correspond to slipping motions if \(r_t\) is small, while more dissipative as \(r_t\) increases.

\subsection{Comparison with an MPCC formulation}

An approach most related with our proposed is the one presented in~\cite{manchester2019}, which is an extension of~\cite{posa2014}.
Therefore, we perform numerical comparisons to understand their differences.
The results are analyzed in terms of the number of iterations and solution time.
Two comparisons are performed for each model, one in a frictionless setting and one with friction.
The mean, standard deviation, and iterations are shown in \cref{tab:mpcc_our_comparison}, where each simulation is repeated \(n\) times.

For the comparison, the relaxation suitable for interior-point methods is utilized~\cite{manchester2019}.
The dynamics for both methods are enforced using the inverse dynamics approach in \cref{subsec:overall_optimization}.
Regarding the slack variables weighing in~\cite{manchester2019}, \(\alpha=1\) is selected. The other aspects of the mathematical program with complementarity constraints remain intact.

\begin{figure}
	\centering
	\includegraphics[width=\linewidth]{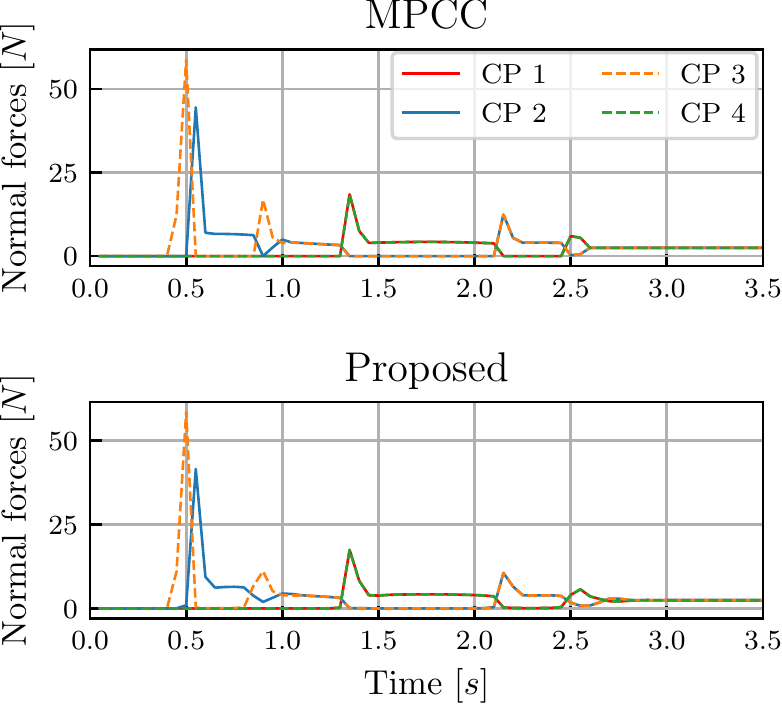}
	\caption{Comparison of the normal impulses at the four contact points (CP) of the brick with friction. The contact impulses at the other four unactivated vertices are zero and not displayed.}\label{fig:vs_mpcc_impulses}
\end{figure}

\subsubsection{Ball model}

For both methods, the friction coefficient is \(\mu = 0.5\), and our model parameters were chosen as \(r_n = 100\nicefrac{N}{m}\) and \(r_t = 1\nicefrac{N}{\nicefrac{m}{s}}\) in the friction case, to match the response with the contact model used by the MPCC\@.
We initialize both methods with zero variables, while we select a time step \(h = 0.1s\) and final time \(t_f = 1s\).

The root-mean-square error (RMSE) for the frictionless case is \(0.0218 N\) (only normal force component), while for the friction case is \(\left(0.0037, 0.007, 0.0218\right) N\); the ball's mass is \(0.2kg\).
Note that tangential forces are generated only at the knot after the contact event and are zero otherwise.
Also, the MPCC method uses a linearized friction cone, whereas we use the full cone model.
The linearized version does not properly capture solutions that lie on the boundary.

\subsubsection{Brick model}

We select \(\mu = 0.6\), \(r_n = 1000\nicefrac{N}{m}\) and \(r_t = 10\nicefrac{N}{\nicefrac{m}{s}}\) for our model's parameters.
We initialize both methods with zero variables, with a time step \(h = 0.05s\) for a horizon of \(t_f = 3.5s\).

In the frictionless case, the number of variables for the MPCC formulation is \(1540\), while our formulation has \(420\).
Both methods have the same number of equalities (\(420\)), while MPCC additionally includes \(2240\) inequalities.
It is also very common to experience plateaus during the iterations with the MPCC approach, while we avoid such a problem by adapting the parameters.
These plateaus are the main reason for the increased number of iterations in the brick frictionless case.

Similarly, the number of variables of the MPCC problem is \(6580\) for the case with friction, while for the proposed method is again \(420\).
The MPCC problem has \(2660\) equality and \(8960\) inequality constraints, while ours has \(420\) equalities and zero inequalities.
The computed normal contact impulses are shown in~\cref{fig:vs_mpcc_impulses}, where the results show only the four activated contact points for clarity.

\begin{figure}
	\centering
	\includegraphics[width=\linewidth]{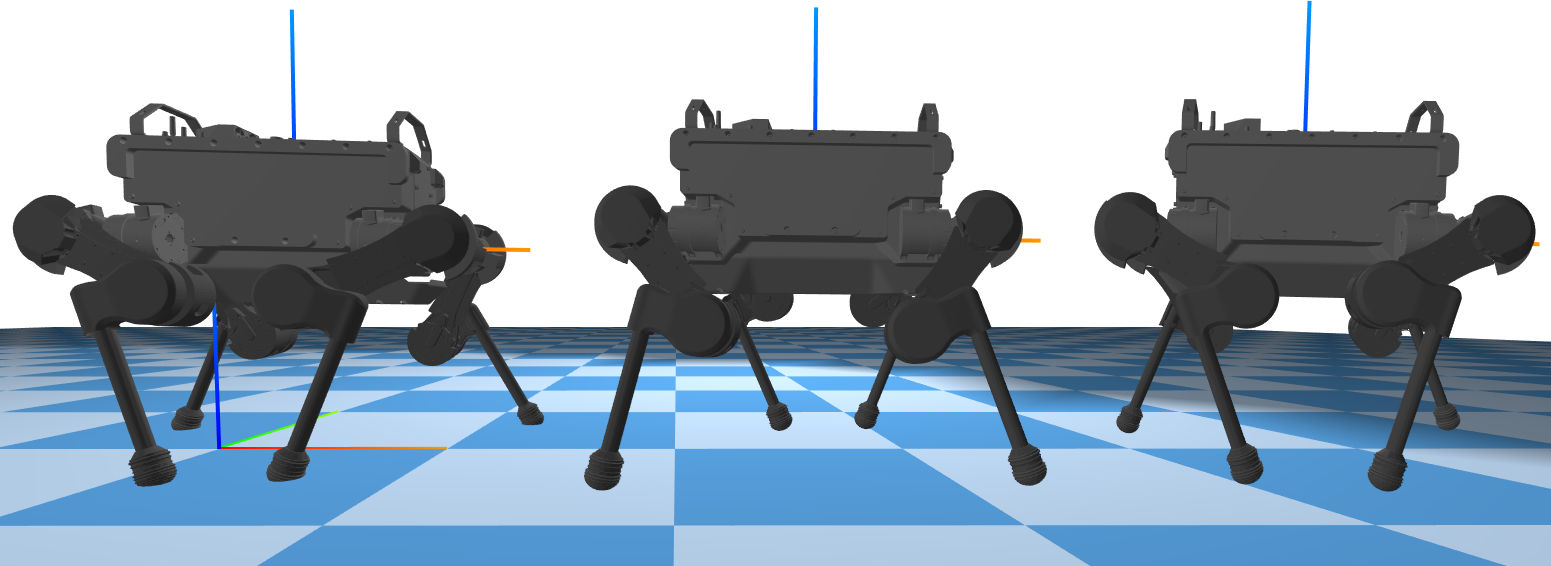}
	\caption{Trotting on hard ground snapshots (left to right).}\label{fig:trot_snapshots}
\end{figure}

\begin{figure*}
	\centering
	\includegraphics[width=\linewidth]{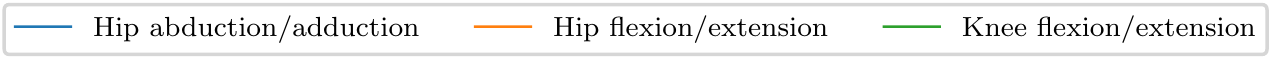}%
	\vspace{3mm}
	\includegraphics[width=0.45\linewidth]{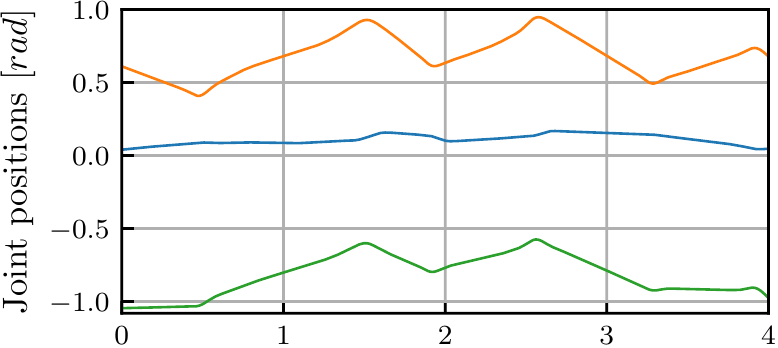}
	\hspace{5mm}
	\includegraphics[width=0.45\linewidth]{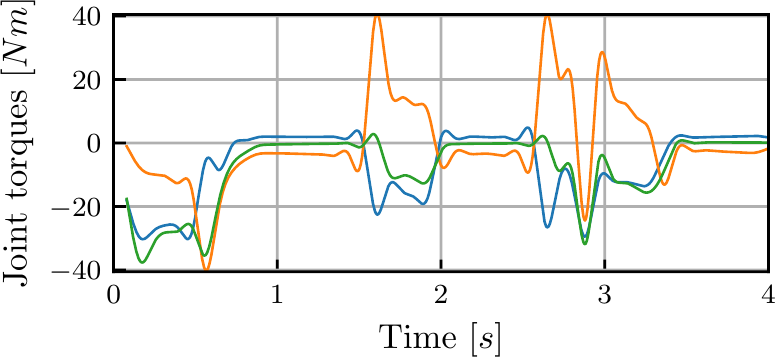}
	\caption{The joint position and torque (40\(Nm\) limit) trajectories of the left front leg (top) and left hind leg (bottom) for the trotting motion on hard ground.}\label{fig:trot_joint_torques}
\end{figure*}

\subsubsection{Overall remarks}

Based on the comparison, there are two characteristics that are advantageous:
\begin{itemize}
	\item The size of the optimization problem is kept minimum as we do not introduce extra variables (\eg slack variables, Lagrange multipliers).
	This is important because general-purpose nonlinear solvers usually demonstrate higher than linear in time complexity since they do not utilize block factorization for solving the problem.
	Also, the optimizer's search space is smaller.

	\item The parameters in our proposed method are few and have direct physical meaning, which makes selection intuitive.
\end{itemize}

For the same problem instantiation, our method generally converges faster and requires a smaller number of steps.
In our experience, choosing parameters that produce similar behaviors is not hard.
For the MPCC method, there are no parameters to select and the performance is fixed for a specific problem instantiation.

Finally, our comparison did not include a cost function for reasons explained before.
This is favorable for the MPCC formulation because selecting \(\alpha\) in general can be challenging.
For actuated systems with multiple solutions, \(\alpha\) needs to be correctly tuned to drive the slack variables to values that sufficiently minimize complementarity violations, without affecting the optimized task.
In our approach, there are no such parameters.
Once appropriate values for \(r_n\) and \(r_t\) are selected to model the environment, they do not change between different tasks.

\subsection{ANYmal trotting on hard and slippery surfaces}

The proposed method is applied to the quadrupedal robot \texttt{ANYmal}.
We use similar gains as in~\cite{neunert2017a} to generate trotting gaits.
The torque limits of the system (\(40Nm\)) are set as inequality constraints.
Since the limits are provided, the torque penalization term in the cost function is decreased while the joint velocity penalization term is increased.

Furthermore, a step size of \(h = 0.08s\) and a horizon of \(t_f = 4s\) is selected, resulting in a problem with \(900\) variables.
By selecting this step size, we aim to demonstrate a positive aspect of our contact model: it is able to handle such large step sizes while not suffering from numerical stiffness.
The step size is \(\times 40\) bigger than the one used in~\cite{neunert2017a} and might not capture the maximum impulse.
Here, we aim to compute a feasible motion plan which can be afterwards interpolated to generate reference trajectories for commaning such a motion on a robot.
Since most robots have rubbers on the feet, a certain level of shock absorption not captured by the model is expected.
Moreover, problems of short impulses are mitigated in case of locomotion on soft ground.

We initialize the optimizer with a nominal standing configuration for the whole duration; the same configuration is set as initial and desired final.
This initialization is used only for the simulation on hard ground.
Using this solution, we initialize the same optimization problem on a slippery surface.
The purpose of this is to demonstrate the motion adaptations due to different environmental properties.
Snapshots of the computed motions are shown in \cref{fig:trot_snapshots}, while in \cref{fig:trot_jump_pos} the position in the x axis for both cases is shown.

\subsubsection{Hard ground}

For this case, we select parameters \(r_n = 20\nicefrac{N}{m}\) and \(r_t = 20\nicefrac{N}{\nicefrac{m}{s}}\).
The diagonal legs step together as expected in a trotting gait, while four steps are taken in total.
The duration of each step is different, as shown in \cref{fig:trot_joint_torques}; a possible reason is due to the requirement of stopping at the end of the motion.
In this case, to start and finish the periodic gait on time, a transient state of fast stepping is necessary.

\subsubsection{Slippery ground}

To simulate a slippery ground, we select \(r_t = 4 \nicefrac{N}{\nicefrac{m}{s}}\).
Compared to the previous solution, there are two notable differences.
First, the ground clearance from the moving legs is significantly smaller.
Since the slipping motions are abrupt, the optimizer tries to keep the contact points closer to the surface for fast activation.
Second, the solution now relies more heavily on the hind legs to push the body forward, while the front legs are mostly used for stabilization.
Similar results are reported in~\cite{carius2019}.

\subsection{ANYmal dynamic jumping}

Next, we compute a jumping motion using the \texttt{ANYmal} model on a hard and a soft ground.
We use a waypoint at the middle of the trajectory for reaching a \(0.8m\) height.
As in the previous case, we specify only the initial and final state, and adapt the gains regarding torque and joint velocity penalization, without a maximum torque inequality constraint.
This is because the specified motion is fast and with the current torque limitations the optimizer would struggle to find a solution that reaches the desired height.

\begin{figure}
	\centering
	\includegraphics[width=\linewidth]{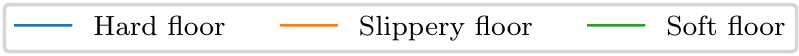}
	\begin{subfigure}[b]{.5\linewidth}
		\centering
		\includegraphics[width=0.9\linewidth]{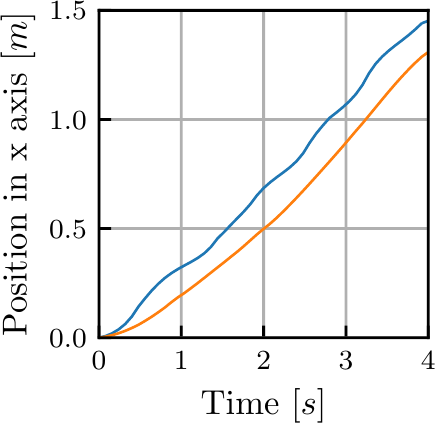}
	\end{subfigure}%
	\begin{subfigure}[b]{.5\linewidth}
		\centering
		\includegraphics[width=0.9\linewidth]{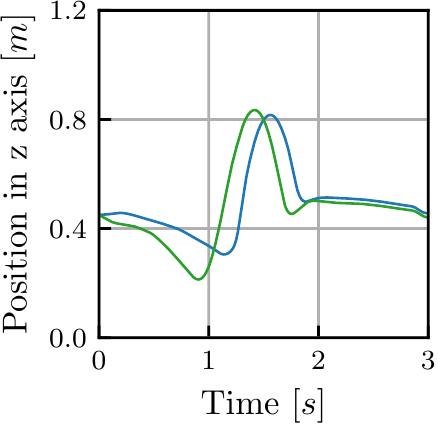}
	\end{subfigure}
	\caption{Body position in the \(x\) axis during trotting on hard and slippery ground, and body position in the \(z\) axis during jumping on hard and soft ground.}\label{fig:trot_jump_pos}
\end{figure}

We select a time horizon of \(t_f = 3s\) and a step size of \(h = 0.06s\), resulting in \(900\) variables.
Again, we use the motion computed on the hard ground to initialize the jumping on the soft ground, aiming to demonstrate the motion adaptation.
Unless specified, we use the same contact model parameters as in the trotting case.
The position of the body in the z axis for both cases is shown in \cref{fig:trot_jump_pos}.

\subsubsection{Hard ground}

The snapshots from the computed solution are shown in \cref{fig:jump_snapshots_hard}.
We identify 6 phases: Lowering the body to prepare for lift-off, the lift-off, reaching the desired height and configuration, touch-down, and transitioning to the desired final state.
Changing the torque penalization weight affects the apex height and time instant that this is reached.

\subsubsection{Soft ground}

\begin{figure}
	\centering
	\begin{subfigure}[b]{.5\linewidth}
		\centering
		\includegraphics[width=0.9\linewidth]{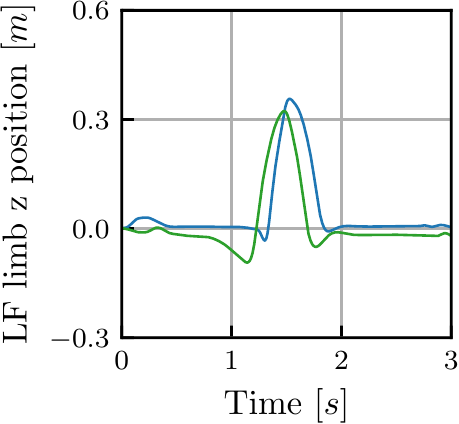}
	\end{subfigure}%
	\begin{subfigure}[b]{.5\linewidth}
		\centering
		\includegraphics[width=0.9\linewidth]{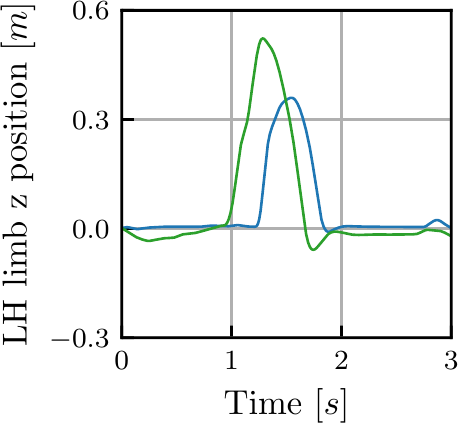}
	\end{subfigure}
	\caption{The foot height during jumping on hard and soft ground for the left front (LF) and left hind (LH) legs.}\label{fig:jump_limb_penetration}
\end{figure}

\begin{figure*}
    \centering
	\begin{minipage}{.48\textwidth}
		\centering
		\vspace{5.5mm}
		\includegraphics[width=0.18\linewidth]{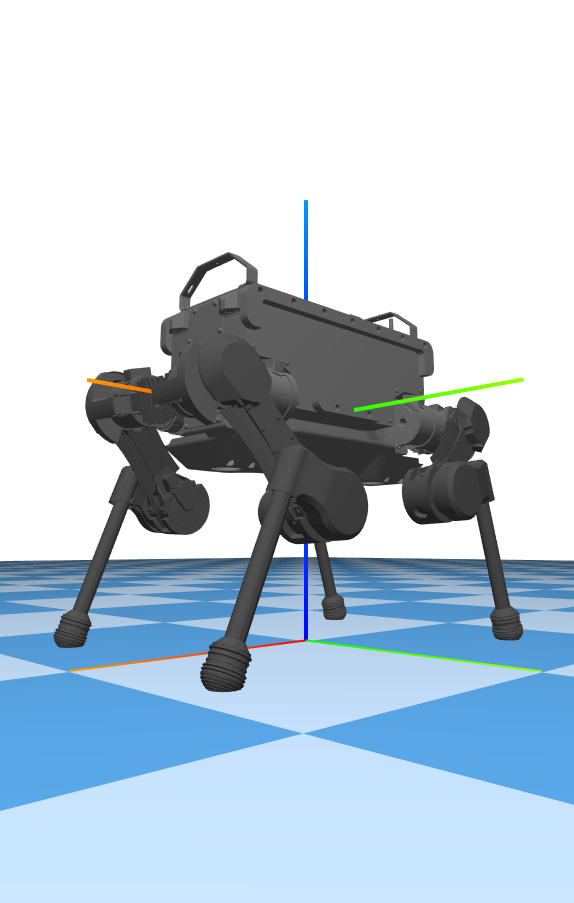}
		\includegraphics[width=0.18\linewidth]{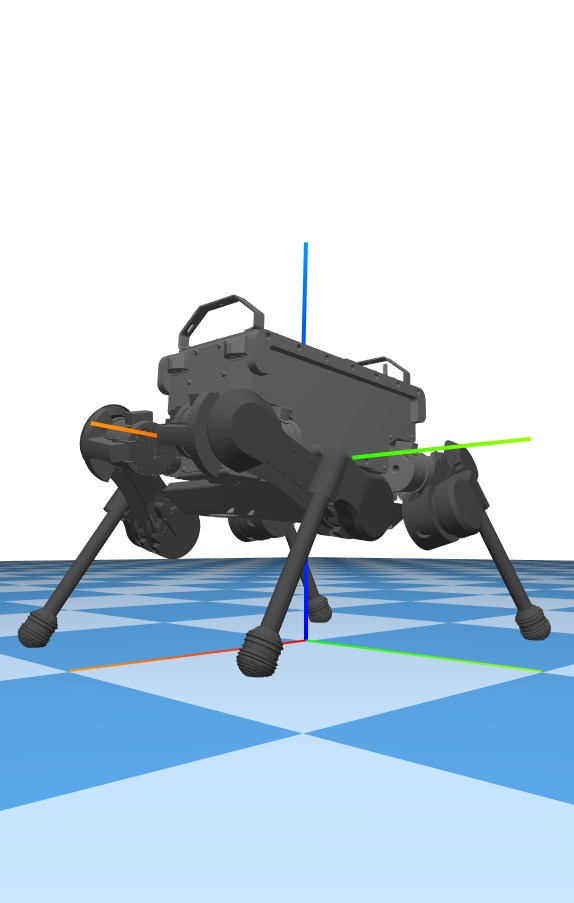}
		\includegraphics[width=0.18\linewidth]{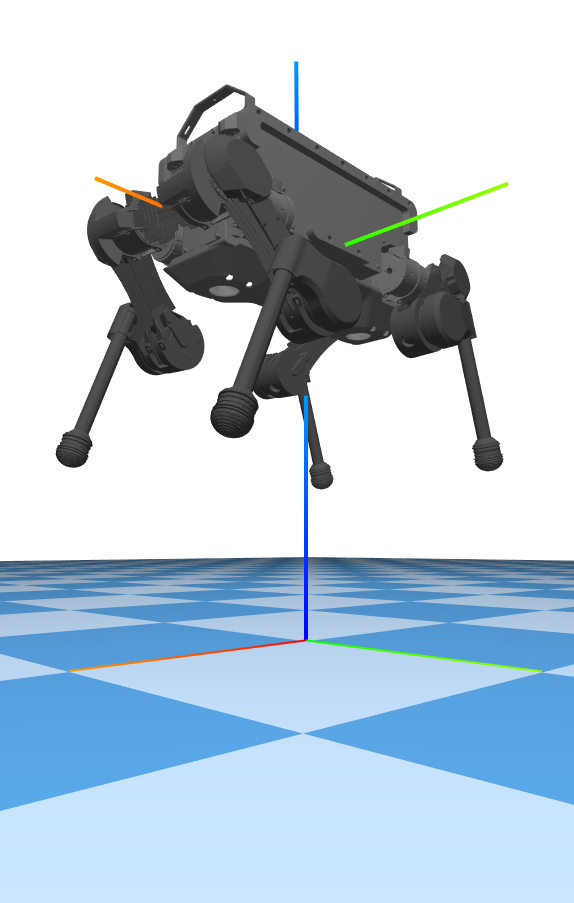}
		\includegraphics[width=0.18\linewidth]{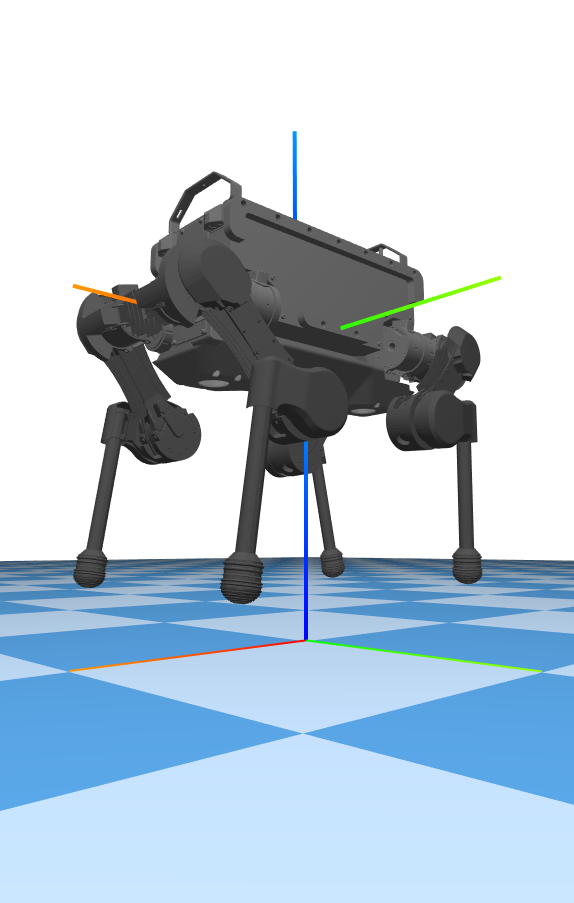}
		\includegraphics[width=0.18\linewidth]{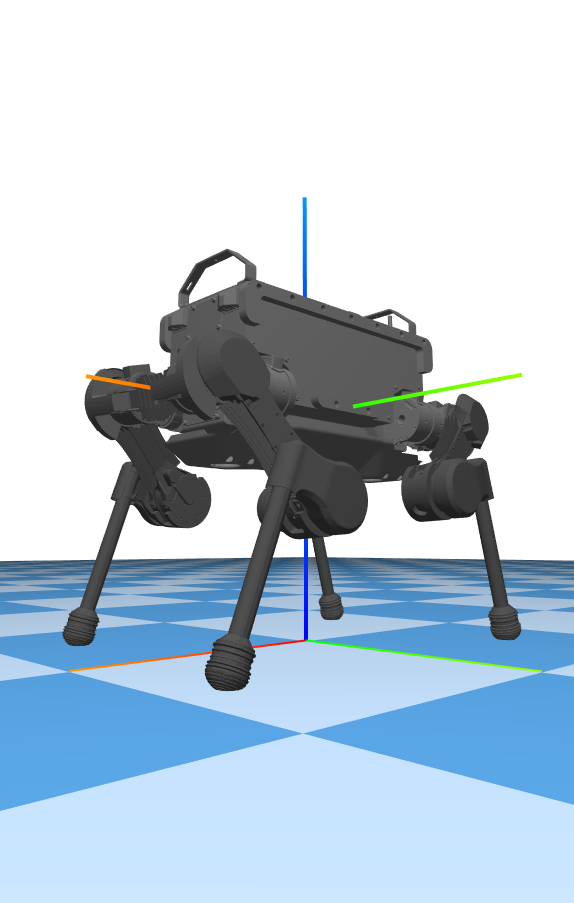}
		\caption{Optimized jumping motion on hard ground, at the same time instances as in the soft ground case.}\label{fig:jump_snapshots_hard}
	\end{minipage}
	\hspace{3mm}
	\begin{minipage}{.48\textwidth}
		\centering
		\vspace{5.5mm}
		\includegraphics[width=0.18\linewidth]{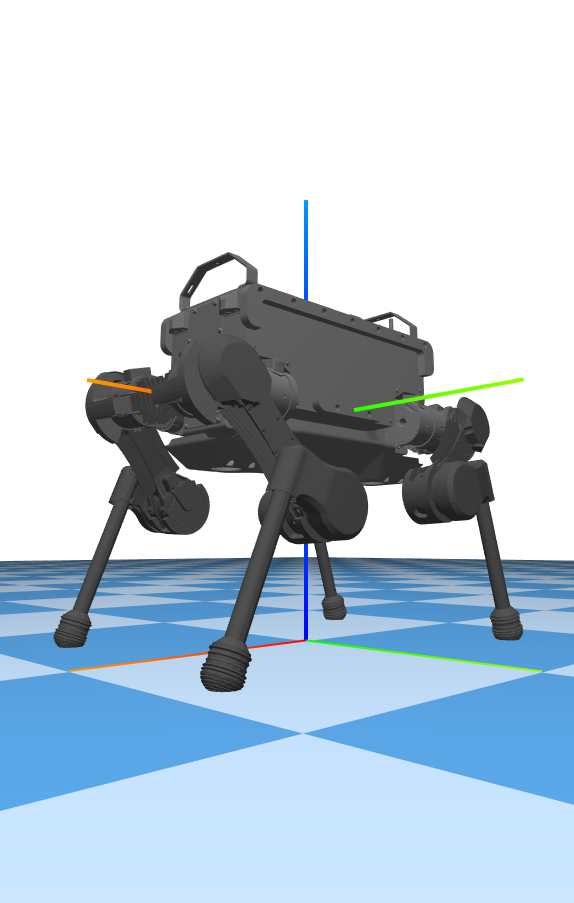}
		\includegraphics[width=0.18\linewidth]{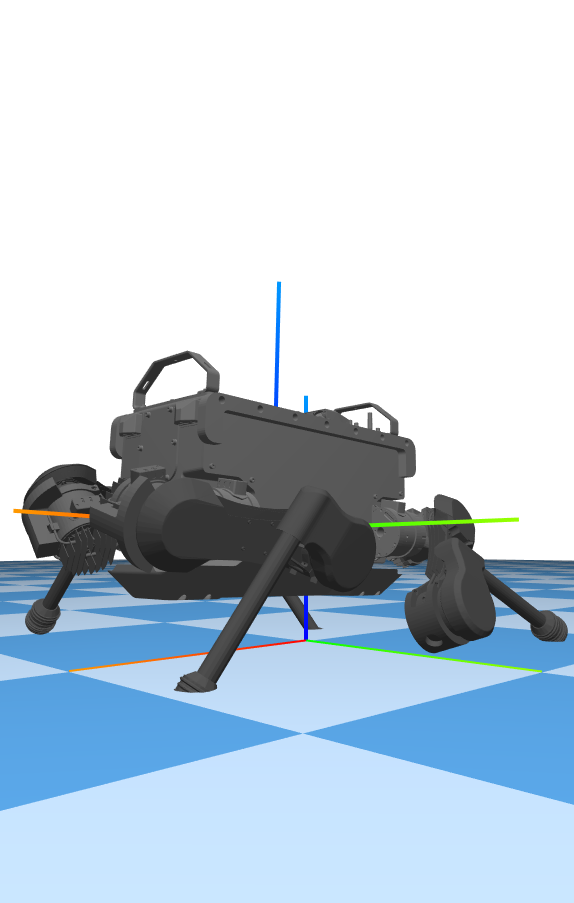}
		\includegraphics[width=0.18\linewidth]{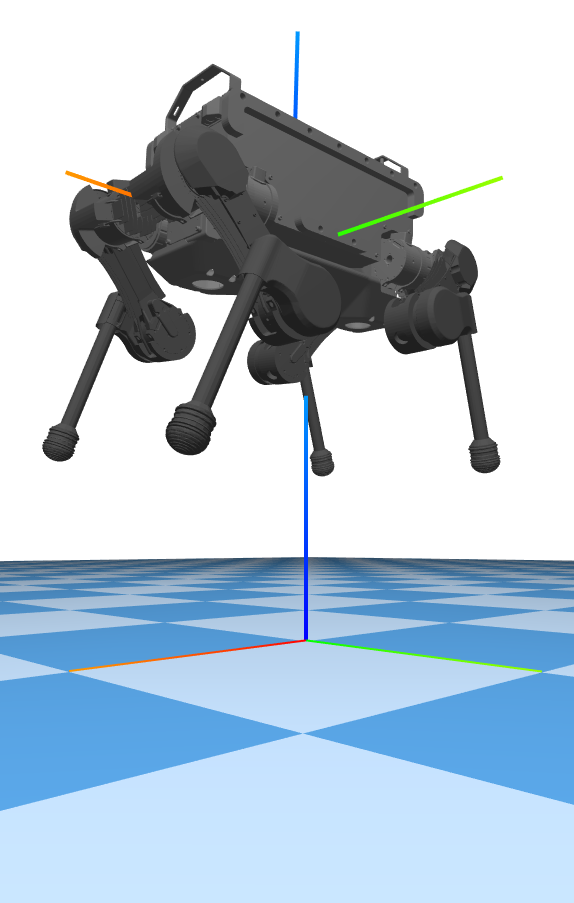}
		\includegraphics[width=0.18\linewidth]{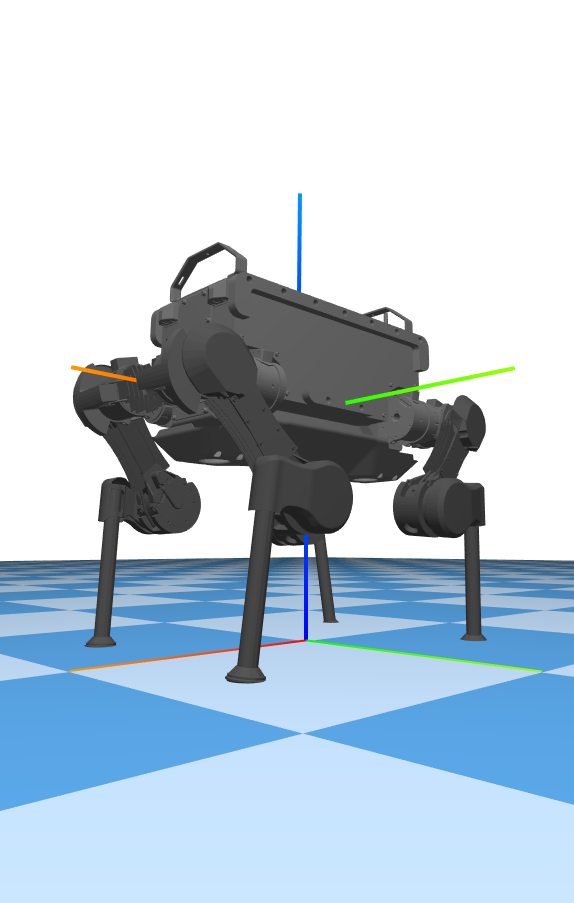}
		\includegraphics[width=0.18\linewidth]{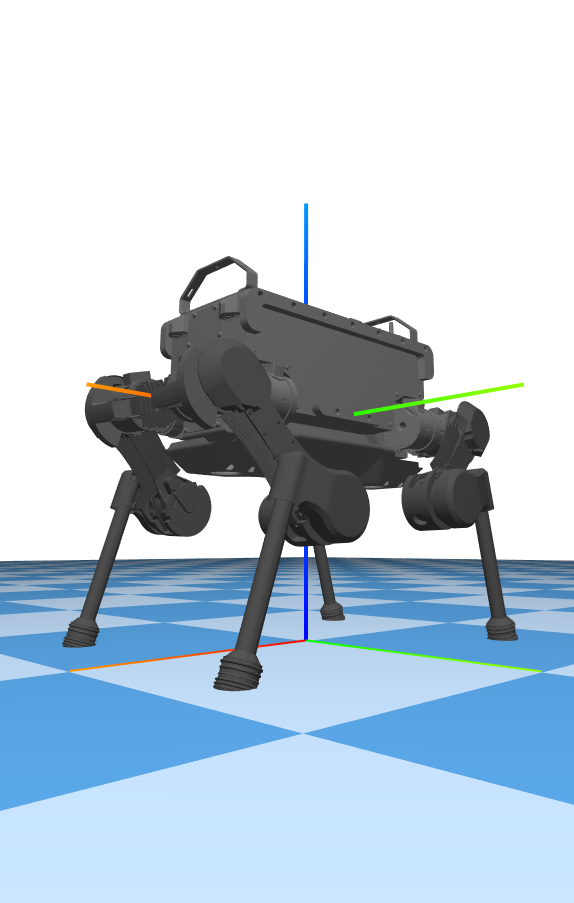}
		\caption{Optimized jumping motion on soft ground, where the feet penetrate the ground surface.}\label{fig:jump_snapshots_soft}
	\end{minipage}
\end{figure*}

For simulating a soft ground, we select \(r_n = 2\nicefrac{N}{m}\).
The penetration of the limb is shown in~\cref{fig:jump_limb_penetration}, while the snapshots of the motion are shown in \cref{fig:jump_snapshots_soft}.
The salient aspects of the motion are the same; differences are identified during the lift-off and touch-down phases.
Specifically, during the lift-off preparation, the body is lowered in a similar manner but the feet penetrate deeper into the soft ground.
Finally, a small oscillation occurrs after touch-down due to the ground's softness and is quickly damped.

\section{Discussion \& Conclusion}\label{sec:conclusion}

This paper proposed a contact model suitable for direct TO formulations, followed by simulation validations with a wide range of models and settings: from simple objects (ball, brick) to a complex multi-body robot model in various locomotion modes (trotting, jumping) and ground conditions (hard, soft, and slippery).
It was demonstrated that the proposed contact-implicit TO method can compute complicated motion plans for multi-contact interactions.
An important feature in the new formulation is an improved, principled contact model which is solved in closed-form and expressed as a function of the state.
Furthermore, this contact model avoids complementarity constraints for its description and automatically satisfies friction cone constraints, while not suffering from problems of energy injections and small step sizes.
Moreover, it is described by two parameters which have intuitive physical interpretation and can be straightforwardly selected.

Nevertheless, there are several aspects worth of discussion and future improvement.
First, the parameters \(r_n\) and \(r_t\) need to be tuned for different robot models or new conditions in the environment's characteristics.
This is a common property among the soft contact models.
Second, the presented method is suitable for solving contact-implicit planning problems in an offline setting, since the computation can not be performed in real-time yet.
The computational time can be improved by taking into account the stage-wise nature of the decision problem.
Lastly, a common feature of TO formulations is that the cost function needs to be specified for each task, \ie the cost function is task-dependent.

The motion adaptations for various ground conditions demonstrate the importance of integrating environmental properties into motion planning.
Future work will focus on systematic identification of parameters \(r_n\) and \(r_t\) and hardware validations.
Another extension of the framework will focus on improving the numerical accuracy by using higher-order integration methods as in~\cite{patel2019a}.

\bibliographystyle{IEEEtran}
\balance%
\bibliography{IEEEabrv,ref}

\end{document}